\documentclass[11pt,a4paper]{article}
\usepackage[hyperref]{sigdial2017}
\usepackage{times}
\usepackage{latexsym}
\usepackage{color}
\usepackage{graphicx}
\usepackage{amsmath}
\usepackage{breakcites}

\usepackage{url}

\aclfinalcopy 
\def\aclpaperid{20} 

\newcommand{\specialcell}[2][c]{%
  \begin{tabular}[#1]{@{}c@{}}#2\end{tabular}}
\newcommand{\whethermath}[1]{\ifmmode{#1}\else{$#1$}\fi}
\newcommand{\uprm}[1]{\whethermath{^{\mbox{\scriptsize #1}}}}

\title{Key-Value Retrieval Networks for Task-Oriented Dialogue}

\author{ Mihail Eric$^{1}$, Lakshmi Krishnan$^{2}$, \\\textbf{Francois Charette}$^{2}$, and \textbf{Christopher D. Manning}$^{1}$\\
{\tt meric@cs.stanford.edu}, {\tt lkrishn7@ford.com} \\
{\tt fcharett@ford.com}, {\tt manning@stanford.edu} \\
Stanford NLP Group$^{1}$ \hspace{0.02in}
Ford Research and Innovation Center$^{2}$}

\date{}

\begin{document}

\maketitle

\begin{abstract}
  Neural task-oriented dialogue systems often struggle to smoothly interface with a knowledge base. In this work, we seek to address this problem by proposing a new neural dialogue agent that is able to effectively sustain grounded, multi-domain discourse through a novel key-value retrieval mechanism. The model is end-to-end differentiable and does not need to explicitly model dialogue state or belief trackers. We also release a new dataset of 3,031 dialogues that are grounded through underlying knowledge bases and span three distinct tasks in the in-car personal assistant space: calendar scheduling, weather information retrieval, and point-of-interest navigation. Our architecture is simultaneously trained on data from all domains and significantly outperforms a competitive rule-based system and other existing neural dialogue architectures on the provided domains according to both automatic and human evaluation metrics.
  
\end{abstract}

\section{Introduction}
  With the success of new speech-based human-computer interfaces, there is a great need for effective task-oriented dialogue agents that can handle everyday tasks such as scheduling events and booking hotels. Current commercial dialogue agents are often brittle pattern-matching systems which are unable to maintain the kind of flexible conversations that people desire. Neural dialogue agents present one of the most promising avenues for leveraging dialogue corpora to build statistical models directly from data by using powerful distributed representations ~\cite{Bordes:16, Wen:16, Dhingra:16}. 

\begin{figure}[ht]
\small
\centering
  \begin{tabular}{|c|c|c|c|c|}
  \hline
  \bf Event & \bf Time & \bf Date & \bf Party & \bf Agenda \\ \hline
  opt. appt. & 10am & Thursday & sister & - \\ 
  dinner & 8pm & the 13th & Ana & - \\
  opt. appt. & 7pm & the 20th & Jeff & - \\ 
  opt. appt. & 4pm & the 13th & Alex & - \\ 
  ... & ... & ... & ... & ... \\
  \hline
  \end{tabular}

  \vspace{3mm}

  \newcommand\D{\emph{D}}

  \begin{tabular}{ll}
    \bf DRIVER: & I need to find out the time and parties \\
    & attending my optometrist appointment.\\
    \bf CAR: & I have 3 appointments scheduled, with Alex, \\
    & your sister, and Jeff. Which are you \\
    & referring to? \\
    \bf DRIVER: & I want to know about the one that Alex is \\ 
    & joining me at \\
    \bf CAR: & That optometrist appointment is at 4 pm.\\
    \bf DRIVER: & Thanks\\
    \bf CAR: & no problem \\
  \end{tabular}

  \caption{Sample dialogue from our dataset. Note some columns and rows from the knowledge base are not included due to space constraints. A dash indicates a missing value.}
\end{figure}

  While this work has been somewhat successful, these task-oriented neural dialogue models suffer from a number of problems: 1) They struggle to effectively reason over and incorporate knowledge base information while still preserving their end-to-end trainability and 2) They often require explicitly modelling user dialogues with belief trackers and dialogue state information, which necessitates additional data annotation and also breaks differentiability. 

  To address some of the modelling issues in previous neural dialogue agents, we introduce a new architecture called the Key-Value Retrieval Network. This model augments existing recurrent network architectures with an attention-based key-value retrieval mechanism over the entries of a knowledge base, which is inspired by recent work on key-value memory networks~\cite{miller-EtAl:2016:EMNLP2016}. By doing so, it is able to learn how to extract useful information from a knowledge base directly from data in an end-to-end fashion, without the need for explicit training of belief or intent trackers as is done in traditional task-oriented dialogue systems. The architecture has no dependence on the specifics of the data domain, learning how to appropriately incorporate world knowledge into its dialogue utterances via attention over the key-value entries of the underlying knowledge base.

  In addition, we introduce and make publicly available a new corpus of 3,031 dialogues spanning three different domain types in the in-car personal assistant space: calendar scheduling, weather information retrieval, and point-of-interest navigation. The dialogues are grounded through knowledge bases. This makes them ideal for building dialogue architectures that seamlessly reason over world knowledge. The multi-domain nature of the dialogues in the corpus also makes this dataset an apt test bed for generalizability of modelling architectures.\footnote{The data is available for download at https://nlp.stanford.edu/blog/a-new-multi-turn-multi-domain-task-oriented-dialogue-dataset/}

  The main contributions of our work are therefore two-fold: 1) We introduce the Key-Value Retrieval Network, a highly performant neural task-oriented dialogue agent that is able to smoothly incorporate information from underlying knowledge bases through a novel key-value retrieval mechanism. Unlike other dialogue agents which only rely on prior dialogue history for generation~\cite{45189,E17-2075}, our architecture is able to access and use database-style information, while still retaining the text generation advantages of recent neural models. By doing so, our model outperforms a competitive rule-based system and other baseline neural models on a number of automatic metrics as well as human evaluation. 2) We release a new publicly-available dialogue corpus across three distinct domains in the in-car personal assistant space that we hope will help further work on task-oriented dialogue agents.

\section{Key-Value Retrieval Networks}
  While recent neural dialogue models have explicitly modelled dialogue state through belief and user intent trackers ~\cite{Wen:16,Dhingra:16,Henderson:14b}, we choose instead to rely on learned neural representations for implicit modelling of dialogue state, forming a truly end-to-end trainable system. Our model starts with an encoder-decoder sequence architecture and is further augmented with an attention-based retrieval mechanism that effectively reasons over a key-value representation of the underlying knowledge base. We describe each component of our model in the subsequent sections.

\subsection{Encoder}
  Given a dialogue between a user (\emph{u}) and a system (\emph{s}), we represent the dialogue utterances as $\{(u_1, s_1), (u_2, s_2), \ldots ,(u_k, s_k) \}$ where $k$ denotes the number of turns in the dialogue. At the $i\uprm{th}$ turn of the dialogue,
we encode the aggregated dialogue context composed of the tokens of $(u_1, s_1, \ldots, s_{i-1}, u_i)$. Letting $x_1, \ldots  , x_m$ denote these tokens,
we first embed these tokens using a trained embedding function $\phi^{emb}$ that maps each token to a fixed-dimensional vector. These mappings are fed into the encoder to produce context-sensitive hidden representations $h_1, \ldots , h_m$, by repeatedly applying the recurrence:
  \begin{align}
      h_i &= \textrm{LSTM}(\phi^{emb}(x_i), h_{i-1})
  \end{align}
  where the recurrence uses a long-short-term memory unit, as described by ~\cite{Hochreiter:97}.

\subsection{Decoder}

The vanilla sequence-to-sequence decoder predicts the tokens of the $i\uprm{th}$ system response $s_i$ by first computing decoder hidden states via the recurrent unit. We denote $\tilde h_1, \ldots, \tilde h_n$ as the hidden states of the decoder and $y_1, \ldots, y_n$ as the output tokens.  We extend this decoder with an attention-based model \cite{Bahdanau:14,Luong:15a}, where, at every time step $t$ of the decoding, an attention score $a_{i}^t$ is computed for each hidden state $h_i$ of the encoder, using the attention mechanism of ~\cite{NIPS2015_5635}. Formally this attention can be described by the following equations:
    \begin{align}
          u_i^t &= w^T \tanh(W_2\tanh(W_1[h_i, \tilde h_t]))) \\
          a_i^t &= \textrm{Softmax}(u_i^t)\\
          \tilde h_t' &= \displaystyle \sum_{i=1}^m a_i^th_i \\
          o_t &= U[\tilde h_t, \tilde h_t'] \\
          y_t &= \textrm{Softmax}(o_t)
    \end{align}

  \noindent where $U$, $W_1$, $W_2$, and $w$ are trainable parameters of the model and $o_t$ represents the logits over the tokens of the output vocabulary $V$. In (2) above, the attention logit on $h_i$ is computed via a two-layer MLP function with a $\tanh$ nonlinearity at the intermediate layers.  During training, the next token $y_t$ is predicted so as to maximize the log-likelihood of the correct output sequence given the input sequence.

  \begin{figure*}[ht]
    \includegraphics[width=\textwidth,height=8cm]{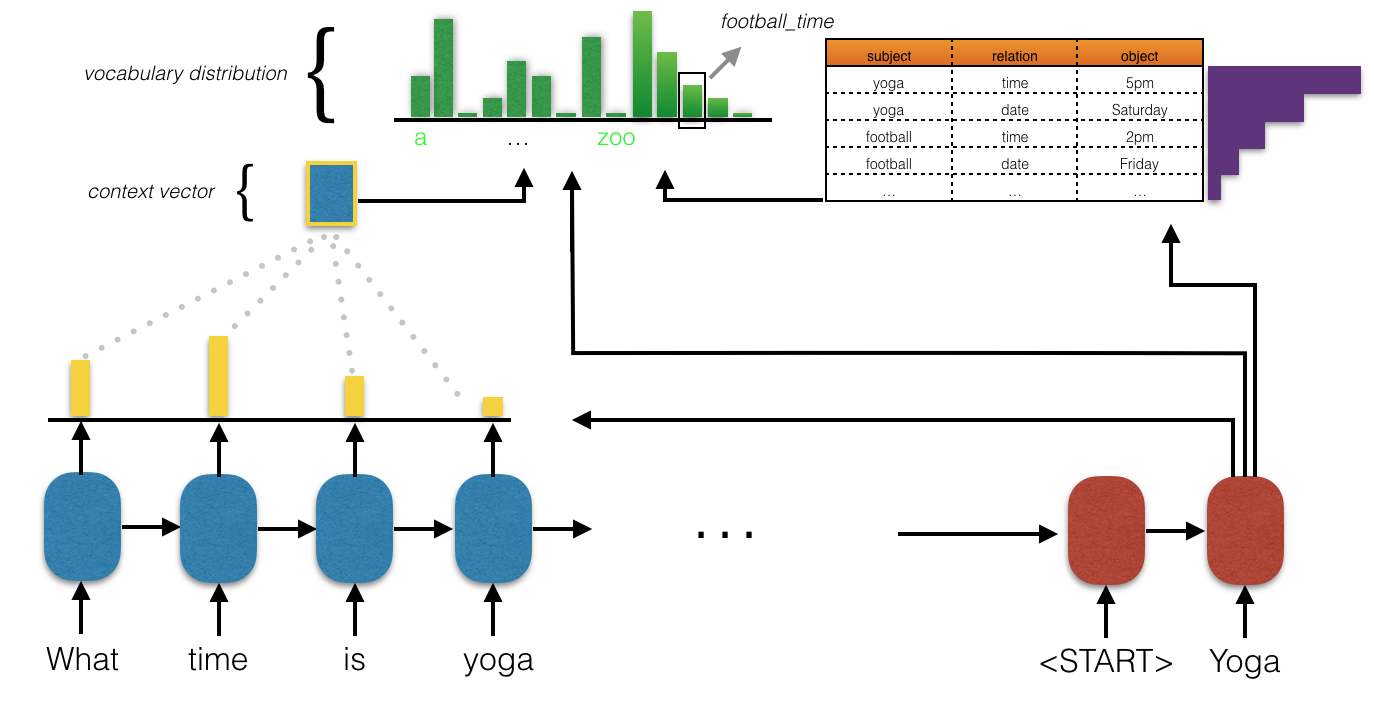}
    \caption{Key-value retrieval network. For each time-step of decoding, the cell state is used to compute an attention over the encoder states and a separate attention over the key of each entry in the KB. The attentions over the encoder are used to generate a context vector which is combined with the cell state to get a distribution over the normal vocabulary. The attentions over the keys of the KB become the logits for their associated values and are separate entries in a now augmented vocabulary that we argmax over.}
  \end{figure*}

\subsection{Key-Value Knowledge Base Retrieval}

  Recently, some neural task-oriented dialogue agents that query underlying knowledge bases (KBs) and extract relevant entities either do the following: 1) create and execute well-formatted API calls to the KB, operations which require intermediate supervision in the form of training slot trackers and which break differentiability ~\cite{Wen:16}, or 2) softly attend to the KB and combine this probability distribution with belief trackers as state input for a reinforcement learning policy ~\cite{Dhingra:16}. We choose to build off the latter approach as it fits nicely into the end-to-end trainable framework of sequence-to-sequence modelling, though we are in a supervised learning setting and we do away with explicit representations of belief trackers or dialogue state. 

  For storing the KB of a given dialogue, we take inspiration from the work of ~\cite{miller-EtAl:2016:EMNLP2016} which found that a key-value structured memory allowed for efficient machine reading of documents. We store every entry of our KB using a \emph{(subject, relation, object)} representation. In our representation a KB entry from the dialogue in Figure 1 such as (\textbf{event}=\emph{dinner}, \textbf{time}=\emph{8pm}, \textbf{date}=\emph{the 13th}, \textbf{party}=\emph{Ana}, \textbf{agenda}=``\emph{-}'') would be normalized into four separate triples of the form (\emph{dinner}, \emph{time}, \emph{8pm}). Every KB has at most 230 normalized triples. This formalism is similar to a neo-Davidsonian or RDF-style representation of events.

  Recent literature has shown that incorporating a copying mechanism into neural architectures improves performance on various sequence-to-sequence tasks~\cite{jia-liang:2016:P16-1,gu-EtAl:2016:P16-1,ling-EtAl:2016:P16-1,gulcehre-EtAl:2016:P16-1,E17-2075}. We build off this intuition in the following way: at every timestep of decoding, we take the decoder hidden state and compute an attention score with the key of each normalized KB entry. For our purposes, the key of an entry corresponds to the sum of the word embeddings of the subject (\emph{meeting}) and relation (\emph{time}). The attention logits then become the logits of the value for that KB entry. For our KB attentions, we replace the embedding of the value with a canonicalized token representation. For example, the value \emph{5pm} is replaced with the canonicalized representation \emph{meeting\_time}. At runtime, if we decode this canonicalized representation token, we convert it into the actual value of the KB entry (\emph{5pm} in our running example) through a KB lookup. Note that this means we are expanding our original output vocabulary to $|V| + n$ where $n$ is the number of separate canonical key representation KB entries. 

  In particular, let $k_j$ denote the word embedding of the key of our $j\uprm{th}$ normalized KB entry. We can now formalize the decoding for our KB attention-based retrieval. Assume that we have $m$ distinct triples in our KB and that we are in the $t\uprm{th}$ timestep of decoding: 
    \begin{align}
          u_j^t &= r^T \tanh(W_2'\tanh(W_1'[k_j, \tilde h_t]))) \\
          o_t &= U[\tilde h_t, \tilde h_t'] + \bar v^t \\
          y_t &= \textrm{Softmax}(o_t)
    \end{align}

    \noindent where $r$, $W_1'$, and $W_2'$ are trainable parameters. In (8) above, $\bar v^t$ is a sparse vector with length $|V| + n$. Within $\bar v^t$, the entry for the value embedding $v_j$ corresponding to the key $k_j$ is equal to the logit score $u_j^t$ on $k_j$. Hence, the $m$ entries of $\bar v^t$ corresponding to the values in the KB are non-zero, whereas the remaining entries corresponding to the original vocabulary tokens are $0$. This sparse vector contains our aggregated KB logit scores which we combine with the original logits to get a modified $o_t$. We then select the argmax token as input to the next timestep. This description seeks to capture the intuition that in response to the query \emph{What time is my meeting}, we want the model to put a high attention weight on the key representation for the (\emph{meeting}, \emph{time}, \emph{5pm}) KB triple, which should then lead the model to favor outputting the value token at the given timestep. We provide a visualization of the Key-Value Retrieval Network in Figure 2.

\section{A Multi-Turn, Multi-Domain Dialogue Dataset}
    In an effort to further work in multi-domain dialogue agents, we built a corpus of multi-turn dialogues in three distinct domains: calendar scheduling, weather information retrieval, and point-of-interest navigation. While these domains are different, they are all relevant to the overarching theme of tasks that users would expect of a sophisticated in-car personal assistant.\\

\subsection{Data Collection}
  The data for the multi-turn dialogues was collected using a Wizard-of-Oz scheme inspired by that of ~\cite{Wen:16}. In our scheme, users had two potential modes they could play: \emph{Driver} and \emph{Car Assistant}. In the \emph{Driver} mode, users were presented with a task that listed certain information they were trying to extract from the \emph{Car Assistant} as well as the dialogue history exchanged between \emph{Driver} and \emph{Car Assistant} up to that point. An example task presented could be: \emph{You want to find what the temperature is like in San Mateo over the next two days}. The \emph{Driver} was then only responsible for contributing a single line of dialogue that appropriately continued the discourse given the prior dialogue history and the task definition.

  Tasks were randomly specified by selecting values (\emph{5pm}, \emph{Saturday}, \emph{San Francisco}, etc.) for three to five slots ({\tt time}, {\tt date}, {\tt location}, etc.), depending on the domain type. Values specified for the slots were chosen according to a uniform distribution from a per-domain candidate set.

  \begin{table*}[ht]
\small
\centering
\begin{tabular}{l|c|c|c}
& \bf Calendar Scheduling & \bf Weather Information Retrieval & \bf POI Navigation \\ \hline
Slot Types & \specialcell{event, time, date,\\ party, room, agenda} &  \specialcell{location, weekly time, \\ temperature, weather attribute} &  \specialcell{POI name, traffic info, \\POI category, address, distance}\\
\hline  
\# Distinct Slot Values & 79 & 65 & 140 \\
\hline
\end{tabular}
\caption{\label{domain-slots} Slots types and number distinct slot values for different domains. POI denotes point-of-interest.}
\end{table*}

  In the \emph{Car Assistant} mode, users were presented with the dialogue history exchanged up to that point in the running dialogue and a private knowledge base known only to the \emph{Car Assistant} with information that could be useful for satisfying the \emph{Driver} query. Examples of knowledge bases could include a calendar of event information, a collection of weekly forecasts for nearby cities, or a collection of nearby points-of-interest with relevant information. The \emph{Car Assistant} was then responsible for using this private information to provide a single utterance that progressed the user-directed dialogues. The \emph{Car Assistant} was also asked to fill in dialogue state information for mentioned slots and values in the dialogue history up to that point. 

  Each private knowledge base had six to seven distinct rows and five to seven attribute types. The private knowledge bases used were generated by uniformly selecting a value for a given attribute type, where each attribute type had a variable number of candidate values. Some knowledge bases intentionally lacked attributes to encourage diversity in discourse.

  \begin{table}[h]
\centering
\small
\begin{tabular}{cc}
\begin{tabular}{|l|r|}
\hline
Training Dialogues & 2,425  \\
Validation Dialogues & 302  \\
Test Dialogues & 304  \\
Calendar Scheduling Dialogues & 1034 \\
Navigation Dialogues & 1000 \\
Weather Dialogues & 997 \\
Avg. \# of Utterances Per Dialogue & 5.25 \\
Avg. \# of Tokens Per Utterance & 9 \\
Vocabulary Size & 1,601 \\
\# of Distinct Entities & 284 \\
\# of Entity (or Slot) Types & 15 \\
\hline
\end{tabular}
\end{tabular}
\caption{Statistics of Dataset.}\label{tab:stats}
\end{table}

 During data collection, some of the dialogues in the calendar scheduling domain did not explicitly require the use of a KB. For example, in a task such as \emph{Set a meeting reminder at 3pm}, we hoped to encourage dialogues that required the \emph{Car Assistant} to execute a task while asking for \emph{Driver} clarification on underspecified information. Roughly half of the scheduling dialogues fell into this category.

  While specifying the attribute types and values in each task presented to the \emph{Driver} allowed us to ground the subject of each dialogue with our desired entities, it would occasionally result in more mechanical discourse exchanges. To encourage more naturalistic, unbiased utterances, we had users record themselves saying commands in response to underspecified visual depictions of an action a car assistant could perform. These commands were transcribed and then inserted as the first exchange in a given dialogue on behalf of the \emph{Driver}. Roughly $\sim$1,500 of the dialogues employed this transcribed audio command first-utterance technique.

  241 unique workers from Amazon Mechanical Turk were anonymously recruited to use the interface we built over a period of about six days. Data statistics are provided in Table 1 and slot types and values are provided in Table 2. A screenshot of the user-facing interfaces for the data collection, as well as a visual used to prompt user recorded commands, are provided in the supplementary material. \\

\section{Related Work}
   Task-oriented agents for spoken dialogue systems have been the subject of extensive research effort. One line of work by ~\cite{Young:13} has tackled the problem using partially observable Markov decision processes and reinforcement learning with carefully designed action spaces, though the number of distinct action states makes this approach often brittle and computationally intractable.

  The recent successes of neural architectures on a number of traditional natural language processing subtasks ~\cite{Bahdanau:14,NIPS2014_5346,NIPS2015_5635} have motivated investigation into dialogue agents that can effectively make use of distributed neural representations for dialogue state management, belief tracking, and response generation. Recent work by ~\cite{Wen:16} has built systems with modularly-connected representation, belief state, and generation components. These models learn to explicitly represent user intent through intermediate supervision, which breaks end-to-end trainability. Other work by ~\cite{Bordes:16,Liu-Perez:16} stores dialogue context in a memory module and repeatedly queries and reasons about this context to select an adequate system response from a set of all candidate responses. 

  Another line of recent work has developed task-oriented models which are amenable to both supervised learning and reinforcement learning and are able to incorporate domain-specific knowledge via explicitly-provided features and model-output restrictions ~\cite{Williams:17}. Our model contrasts with these works in that training is done in a strictly supervised fashion via a per utterance token generative process, and the model does not need dialogue state trackers, relying instead on latent neural embeddings for accurate system response generation.

  Research in task-oriented dialogue often struggles with a lack of standard, publicly available datasets. Several classical corpora have consisted of moderately-sized collections of dialogues related to travel-booking ~\cite{Hemphill:90,Bennett:02}. Another well-known corpus is derived from a series of competitions on the task of dialogue-state tracking ~\cite{Williams:13}. While the competitions were designed to test systems for state tracking, recent work has chosen to repurpose this data by only using the transcripts of dialogues without state annotation for developing systems ~\cite{Bordes:16,Williams:17}. More recently, Maluuba has released a dataset of hotel and travel-booking dialogues collected in a Wizard-of-Oz Scheme with elaborate semantic frames annotated~\cite{El-Asri:16}. This dataset aims to encourage research in non-linear decision-making processes that are present in task-oriented dialogues.

\section{Experiments}
In this section we first introduce the details of the experiments and then present results from both automatic and human evaluation.

\begin{table*}[ht]
\centering
\small
\begin{tabular}{@{}llccccccc@{}}
 {\bf Model} & {\bf BLEU} & {\bf Ent. F$_1$} & {\bf Scheduling Ent. F$_1$} & {\bf Weather Ent. F$_1$} & {\bf Navigation Ent. F$_1$} \\
\hline 
Rule-Based & 6.6 & 43.8  & 61.3 & 39.5 & 40.4\\
Copy Net &   11.0 & 37.0 & 28.1 & \textbf{50.1} & 28.4 \\
Attn. Seq2Seq & 10.2 & 30.0 & 30.0 & 42.4 & 17.9 \\
KV Retrieval Net (no enc. attn.) & 10.8 & 40.9 & 59.5 & 35.6 & 36.6 & \\
KV Retrieval Net & \textbf{13.2} & \textbf{48.0} & \textbf{62.9} & 47.0 & \textbf{41.3} \\
\hline
\emph{Human Performance} & 13.5 & 60.7 & 64.3 & 61.6 & 55.2 \\
\hline
\end{tabular}
\caption{Evaluation on our test data. Bold values indicate best model performance. We provide both an aggregated F$_1$ score as well as domain-specific F$_1$ scores. Attn. Seq2Seq refers to a sequence-to-sequence model with encoder attention. KV Retrieval Net (no enc. attn.) refers to our new model with no encoder attention context vector computed during decoding.}\label{tab:results}
\end{table*}

\subsection{Details}
For our experiments, we divided the dialogues into train/validation/test sets using a 0.8/0.1/0.1 data split and ensured that each domain type was equally represented in each of the splits. 

To reduce lexical variability, in a pre-processing step, we map the variant surface expression of entities to a canonical form using named entity recognition and linking. For example, the surface form \emph{20 Main Street} is mapped to \emph{Pizza My Heart address}. During inference, our model outputs the canonical forms of the entities, and so we realize their surface forms by running the system output through an inverse lexicon. The inverse lexicon converts the entities back to their surface forms by sampling from a multinomial distribution with parameters of the distribution equal to the frequency count of a given surface form for an entity as observed in the training and validation data. Note that for the purposes of computing our evaluation metrics, we operate on the canonicalized forms, so that any non-deterministic variability in surface form realization does not affect the computed metrics.

\subsection{Hyperparameters}

We trained using a cross-entropy loss and the Adam optimizer ~\cite{Kingma:15} with learning rates sampled from the interval $[10^{-4}, 10^{-3}]$. We applied dropout ~\cite{Hinton:12} as a regularizer to the input and output of the LSTM. We also added an $l_2$ regularization penalty on the weights of the model. We identified hyperparameters by random search, evaluating on the held-out validation subset of the data. Dropout keep rates were sampled from $[0.8, 0.9]$ and the $l_2$ coefficient was sampled from $[3\cdot 10^{-6},10^{-5}]$. We used word embeddings, hidden layer, and cell sizes with size 200.  We applied gradient clipping with a clip-value of 10 to avoid gradient explosions during training.
The attention, output parameters, word embeddings, and LSTM weights were randomly initialized from a uniform unit-scaled distribution in the style of ~\cite{Sussillo:15}. We also added a bias of 1 to the LSTM cell forget gate in the style of ~\cite{Pham:14}. 

\subsection{Baseline Models}
We provide several baseline models for comparing performance of the Key-Value Retrieval Network: 
\begin{itemize}
  \item \textbf{Rule-Based Model}: This model is a traditional rule-based system with modular dialogue state trackers, KB query, and natural language generation components. It first does an extensive domain-dependent keyword search in the user utterances to detect intent. The user utterances are also provided to a lexicon to extract any entities mentioned. Collectively, this information forms the dialogue state up to a given point in the dialogue. This dialogue state is used to query the KB as appropriate, and the returned KB values are used to fill in predefined template system responses.

  \item \textbf{Copy-Augmented Sequence-to-Sequence Network}: This model is derived from the work of ~\cite{E17-2075}. It augments a sequence-to-sequence architecture with encoder attention, with an additional attention-based hard-copy mechanism over the KB entities mentioned in the encoder context. This model does not explicitly incorporate information from the underlying KB and instead relies solely on dialogue history for system response generation. Unlike the best performing model of ~\cite{E17-2075}, we do not enhance the inputs to the encoder with additional entity type features, as we found that the model performed worse on our data with this added mechanism. We choose this model for comparison as it is also end-to-end trainable and implicitly models dialogue state through learned neural representations, putting it in the same class of dialogue models as our key-value retrieval net. This model has also been shown to be a competitive task-oriented dialogue baseline that can accurately interpret user input and act on this input through latent distributed representation. We refer to this model as Copy Net in the results tables.

\end{itemize}

\subsection{Automatic Evaluation}

\subsubsection{Metrics}
Though prior work has shown that automatic evaluation metrics often correlate poorly with human assessments of dialogue agents ~\cite{liu-EtAl:2016:EMNLP20163}, we report a number of automatic metrics in Table 3. These metrics are provided for coarse-grained evaluation of dialogue response quality:

\begin{itemize}
  \item \textbf{BLEU}: We use the BLEU metric, commonly employed in evaluating machine translation systems ~\cite{papineni-EtAl:2002:ACL}, which has also been used in past literature for evaluating dialogue systems both of the chatbot and task-oriented variety~\cite{Ritter:11a,li-EtAl:2016:N16-11,Wen:16}. While work by ~\cite{liu-EtAl:2016:EMNLP20163} has demonstrated that n-gram based evaluation metrics such as BLEU and METEOR do not correlate well with human performance on non-task-oriented dialogue datasets, recently ~\cite{Sharma:17} have shown that these metrics can show comparatively stronger correlation with human assessment on task-oriented datasets. We, therefore, calculate average BLEU score over all responses generated by the system, and primarily report these scores to gauge our model's ability to accurately generate the language patterns seen in our data.  

  \item \textbf{Entity F$_1$}: Each human Turker's \emph{Car Assistant} response in the test data defines a gold set of entities. To compute an entity F$_1$, we micro-average over the entire set of system dialogue responses and use the entities in their canonicalized forms. This metric evaluates the model's ability to generate relevant entities from the underlying knowledge base and to capture the semantics of the user-initiated dialogue flow. Given that our test set contains dialogues from all three domains, we compute a per-domain entity F$_1$ as well as an aggregated dataset entity F$_1$. We note that other work on task-oriented dialogue by ~\cite{Wen:16,Henderson:14} have reported the slot-tracking accuracy of their systems, which is a similar but perhaps more informative and fine-grained notion of a system's ability to capture user semantics. Because our model does not have provisions for slot-tracking by design, we are unable to report such a metric and hence report our entity F$_1$. 

\end{itemize}

\subsubsection{Results}

We see that of our baseline models, Copy Net has the lowest aggregate entity F$_1$ performance. Though it has the highest model entity F$_1$ for the weather domain dialogues, it performs very poorly in the other domains, indicating its inability to generalize well to multiple dialogue domains and to accurately integrate relevant entities into its responses. Copy Net does, however, have the second highest BLEU score, which is not surprising given that the model is a powerful extension to the sequence-to-sequence modelling class, which is known to have very robust language modelling capabilities. 

Our rule-based model has the lowest BLEU score, which is a consequence of the fact that the naturalness of the system output is very limited by the number of diverse and distinct response templates we manually provided. This is a common issue with heuristic dialogue agents and one that could be partially alleviated through a larger collection of lexically rich response templates. However, the rule-based system has a very competitive aggregate entity F$_1$. This is because it was designed to accurately parse the semantics of user utterances and query the underlying KB of the dialogue, through manually-provided heuristics. 

 As precursors to our key-value retrieval net, we first report results of a model that does not compute an attention over the KB (referred to as Attn. Seq2Seq) and show that without computing attention over the KB, the model performs poorly in entity F$_1$ as its output is agnostic to the world state represented in the KB. Note that this model is effectively a sequence-to-sequence model with encoder attention. If we include an attention over the KB but do not compute an encoder attention (referred to as KV Retrieval Net no enc. attn.), the entity F$_1$ increases drastically, showing that the model is able to incorporate relevant entities from the KB. Finally, we combine these two attention mechanisms to get our final key-value retrieval net. Our proposed key-value retrieval net has the highest modelling performance in BLEU, aggregate entity F$_1$, and entity F$_1$ for the scheduling and navigation domains. It outperforms the rule-based aggregate entity F$_1$ by $4.2\%$ and outperforms the Copy Net BLEU score by $2.2$ points as well as its entity F$_1$ by $11\%$. These salient gains are noteworthy because our model is able to achieve them by learning its latent representationts directly from data, without the need for heuristics or manual labelling.

We also report human performance on the provided metrics. These scores were computed by taking the dialogues of the test set and having a second distinct batch of Amazon Mechanical Turk workers provide system responses given prior dialogue context. This, in effect, functions as an interannotator agreement score and sets a human upper bound on model performance. We see that there is a sizable gap between human performance on entity F$_1$ and that of our key-value retrieval net ($\sim12.7\%$), though our model is on par with human performance in BLEU score.

\subsection{Human Evaluation}
  We randomly generated 120 distinct scenarios across the three dialogue domains, where a scenario is defined by an underlying KB as well as a user goal for the dialogue (e.g. \emph{find the nearest gas station, avoiding heavy traffic}). We then paired Amazon Mechanical Turkers with one of our systems in a real-time chat environment, where each Turker played the role of the \emph{Driver}. We evaluated the rule-based model, Copy Net, and key-value retrieval network on each of the 120 scenarios. We also paired a Turker with another Turker for each of the scenarios, in order to get evaluations of human performance. At the end of the chat, the Turker was asked to judge the quality of their partner according to fluency, cooperativeness, and humanlikeness on a scale from 1 to 5. The average scores per pairing are reported in Table 4. In a separate experiment, we also had Turkers evaluate the outputs of the systems on 80 randomly selected dialogues from the test split of our dataset. Those outputs were evaluated according to correctness, appropriateness, and humanlikeness of the responses, and the scores are reported in Table 5.

  We see that on real-time dialogues the key-value retrieval network outperforms the baseline models on all of the metrics, with especially sizeable performance gains over the Copy Net which is the only other recurrent neural model evaluated. We also see that human performance on this assessment sets the upper bound on scores, as expected. The results on human evaluation of test outputs show that the rule-based model provides the most correct system responses, the KV network provides the most appropriate responses, and the Copy Net gives the most humanlike responses by small margins. We should note, however, that the second regime for human evaluation is more unrealistic because it involves providing a dialogue context that is directly sampled from our dataset, whereas the first regime of real-time dialogues measures the models' abilities to adapt to new and noisier user input. This suggests that the first set of results are more meaningful and representative for assessing overall model efficacy. 

  Examples of dialogues conducted between our model and Turkers are included in Figure 3. Particularly noteworthy is our model's ability to seamlessly integrate world information from the underlying KBs in the respective dialogues, while still producing very naturalistic utterances. The model is able to do this effectively across multiple domains.

  \begin{figure}[ht]
\small
\centering
  \begin{tabular}{|c|c|c|c|c|}
  \hline
  \bf POI& \bf Category & \bf Traffic Info& \bf ... \\ \hline
  Civic Center & parking garage & car collision & ... \\ 
  Valero & gas station  & road block  & ... \\
  Webster Garage & parking garage & car collision  & ... \\ 
  Trader Joes & grocery store & heavy & ... \\ 
  Mandarin Rts. & chinese rest. & moderate  & ... \\
  ... & ... & ... & ...  \\
  \hline
  \end{tabular}

  \vspace{3mm}

  \newcommand\D{\emph{D}}

  \begin{tabular}{ll}
    \bf DRIVER: & I am looking for a gas station near me with \\
    &  the shortest route from me.\\
    \bf CAR: & \emph{the closest gas station is valero but, but there} \\
    & \emph{is a road block nearby.} \\
    \bf DRIVER: & What is the next nearest gas station?\\ 
    \bf CAR: & \emph{valero is the only gas station that i can find}\\
    \bf DRIVER: & Thanks\\
    \bf CAR: & \emph{here to serve.}\\
  \end{tabular}

  \vspace{8mm}

\small
\centering
  \begin{tabular}{|c|c|c|c|c|}
  \hline
  \bf Event & \bf Time & \bf Date & \bf Party & \bf Agenda \\ \hline
  swimming act. & 3pm & the 11th & sister & - \\ 
  dinner & 7pm & the 3rd & mother & - \\
  football & 2pm & the 20th & mother & - \\ 
  lab appt. & 10am & the 17th & Jeff & - \\ 
  ... & ... & ... & ... & ... \\
  \hline
  \end{tabular}

  \vspace{6mm}

  \begin{tabular}{ll}
    \bf DRIVER: & car when is gonna be my next swimming \\
    & activity? \\
    \bf CAR: & \emph{your next swimming activity is on the 11th}\\
    & \emph{at 3pm.} \\
    \bf DRIVER: & who is gonna attend with me?\\ 
    \bf CAR: & \emph{your swimming activity is on the 11th, one at} \\
    & \emph{3pm with sister}\\
    \bf DRIVER: & thank you!\\
    \bf CAR: & \emph{you are welcome}\\
  \end{tabular}

  \caption{Sample dialogues from our human evaluation, conducted between a Turker and the KV Retrieval Net. In the dialogue, our model responses are italicized. Additional dialogues are included in the supplementary material.}
\end{figure}

\begin{table}
\centering
\small
\begin{tabular}{l|ccc}
{\bf Model} & {\bf Fluency} & {\bf Cooperative} & {\bf Humanlike} \\
\hline
Rule-based & 3.20 & 3.39 & 2.87 \\
Copy Net & 2.33 & 2.42 & 2.02\\
KV Ret. Net & \textbf{3.36} & \textbf{3.43} & \textbf{3.09} \\
\hline
\emph{Human} & 3.98 &  3.95 & 4.04 \\
\hline
\end{tabular}
\caption{Human evaluation results on realtime dialogues.}\label{tab:human-eval}
\end{table}

\begin{table}
\centering
\small
\begin{tabular}{l|ccc}
{\bf Model} & {\bf Correct} & {\bf Appropriate} & {\bf Humanlike} \\
\hline
Rule-based & \textbf{3.96} & 3.57 & 3.28 \\
Copy Net &  3.52 & 3.63 & \textbf{3.56}\\
KV Ret. Net &  3.70 & \textbf{3.64} & 3.50 \\
\hline
\end{tabular}
\caption{Human evaluation of system outputs on test set.}\label{tab:human-eval-test}
\end{table}

\section{Conclusion and Future Work}
  In this work, we have presented a novel neural task-oriented dialogue model that is able to sustain grounded discourse across a variety of domains by retrieving world knowledge represented in knowledge bases. It smoothly incorporates this world knowledge into natural-sounding system responses in an end-to-end trainable fashion, without the need to explicitly model dialogue state. Our model outperforms competitive heuristic and neural baselines on both automatic and human evaluation metrics. In addition, we have introduced a publicly available dialogue dataset across three domains in the in-car personal assistant space that we hope will help the data scarcity issue present in task-oriented dialogue research. 

  Future work will address closing the margin between the Key-Value Retrieval Network and human performance on the various metrics. This will include developing new methods for robust handling of joint KB attributes as well as usage of the KB that requires more pragmatic understanding of the world via notions such as temporal reasoning.

\section*{Acknowledgments}

The authors wish to thank He He, Peng Qi, Urvashi Khandelwal, and Reid Pryzant for their valuable feedback and insights. We gratefully acknowledge the funding of the Ford Research and Innovation Center, under Grant No. 124344.

\bibliography{meric_sigdial2017}
\bibliographystyle{sigdial_natbib}

\appendix

\begin{figure*}
\small
\centering
  \begin{tabular}{|c|c|c|c|c|}
  \hline
  \bf POI& \bf Category & \bf Address & \bf Traffic Info& \bf Distance \\ \hline
  Dish Parking & parking garage & 550 Alester Ave & no traffic & 3 miles \\ 
  Cafe Venetia & coffee or tea place & 329 El Camino Real & road block nearby & 4 miles \\
  Mandarin Roots & chinese restaurant & 271 Springer St & car collision nearby & 3 miles \\ 
  Dominos & pizza restaurant & 776 Arastradero Rd & moderate traffic & 6 miles \\ 
  Travelers Lodge & rest stop & 333 Arbol Dr & moderate traffic & 4 miles \\
  \hline
  \end{tabular}

  \vspace{3mm}

  \newcommand\D{\emph{D}}

  \begin{tabular}{ll}
    \bf DRIVER: & What is the nearest parking garage within 4 miles\\
    &  that has no traffic.\\
    \bf CAR: & \emph{the nearest is dish parking at 550 alester ave, 3 miles away} \\
    \bf DRIVER: & Does it have any traffic?\\ 
    \bf CAR: & \emph{there is no traffic on the route to dish parking}\\
    \bf DRIVER: & Thanks\\
    \bf CAR: & \emph{you're welcome!}\\
  \end{tabular}

  \vspace{6mm}

  \small
\centering
  \begin{tabular}{|c|c|c|c|c|c|}
  \hline
  \bf Event & \bf Time & \bf Date & \bf Party & \bf Agenda & \bf Location \\ \hline
  dinner & 5pm & the 4th & brother & - & -\\ 
  meeting & 1pm & the 18th & vice president & discuss company picnic & room 100 \\ 
  conference & 11am & the 1st & HR & go over budget & room 50 \\ 
  meeting & 4pm & the 9th& HR & onboard new members & room 100\\ 
  lab appt. & 6pm & the 14th & Jon & - & - \\  
  dinner & 6pm & the 1st & Jon & - & -\\
  \hline
  \end{tabular}

  \vspace{3mm}

  \begin{tabular}{ll}
    \bf DRIVER: & When is my dinner going to be and who will be attending\\
    \bf CAR: & \emph{your next dinner is on the 1st at 6 pm with your jon} \\
    \bf DRIVER: & Great, thanks.\\ 
    \bf CAR: & \emph{you're welcome!}\\
  \end{tabular}

  \vspace{6mm}

  \small
\centering
  \begin{tabular}{|c|c|c|c|c|}
  \hline
  \bf POI& \bf Category & \bf Address & \bf Traffic Info& \bf Distance \\ \hline
  Dominos & pizza restaurant & 776 Arastradero Rd & moderate traffic & 7 miles \\ 
  Stanford Express Care & hospital & 214 El Camino Real & car collision nearby & 1 miles \\
  Travelers Lodge & rest stop & 333 Arbol Dr & no traffic & 4 miles \\ 
  Coupa & coffee or tea place & 394 Van Ness Ave & moderate traffic & 7 miles \\ 
  The Clement Hotel & rest stop & 657 Ames Ave & road block nearby & 6 miles \\
  5672 barringer st. & certain address & 5672 barringer st. & car collision nearby & 4 miles \\
  \hline
  \end{tabular}

  \vspace{3mm}

  \begin{tabular}{ll}
    \bf DRIVER: & Hi, I'd like to find a coffee place near me\\
    \bf CAR: & \emph{we're 7 miles away from coupa but there is a moderate traffic} \\
    \bf DRIVER: & What's the address?\\ 
    \bf CAR: & \emph{the address is 394 van ness ave, i sent it on your screen}\\
    \bf DRIVER: & Great. Can you find me the fastest route there?\\
    \bf CAR: & \emph{the fastest route is for coupa.}\\
    \bf DRIVER: & Ok, so coupa is the closest place and you have the fastest route there\\
    \bf CAR: & \emph{the fastest route has moderate traffic and there is a moderate traffic on the route}\\
    & \emph{do you want to go there?}\\
    \bf DRIVER: & yes \\
    \bf CAR: & \emph{i sent the info on your screen, drive carefully!} \\
    \bf DRIVER: & Thanks \\
    \bf CAR: & \emph{you're welcome, have a great day.}
  \end{tabular}

  \vspace{6mm}

  \small
\centering
  \begin{tabular}{|l|c|c|c|c|c|c|}
  \hline
  & \bf \emph{monday} & \bf tuesday & \bf wednesday & \bf thursday & \bf friday & ...\\ \hline
  los angeles& hail, 60-70f& foggy, 30-40f & cloudy, 20-30f & stormy 30-40f & drizzle, 30-50f & ...\\ 
  durham & rain, 80-90f & cloudy, 50-70f & misty, 40-60f & dry, 80-100f & rain, 50-70f & ...\\
  boston& overcast, 30-50f & misty, 40-60f & foggy, 40-50f & foggy, 40-50f & clear skies 20-30f & ...\\ 
  seattle & warm, 60-70f & foggy, 30-40f & cloudy, 20-30f & stormy, 30-40f & drizzle, 30-50f & ...\\ 
  fresno& clear skies, 40-60f & drizzle, 70-80f & humid, 70-90f & stormy, 60-70f & cloudy, 90-100f & ...\\
  san francisco & foggy, 40-60f & hot, 90-100f & humid, 50-70f & rain, 80-100f & misty, 20-40f & ...\\
  alhambra & dew, 30-40f & dry, 30-50f & dry, 50-60f & humid, 80-90f & snow, 60-80f & ...\\
  \hline
  \end{tabular}

  \vspace{3mm}

  \begin{tabular}{ll}
    \bf DRIVER: & how is the weather\\
    \bf CAR: & \emph{what city would you like the weather for?} \\
    \bf DRIVER: & is it foggy in seattle?\\ 
    \bf CAR: & \emph{it will not be foggy in seattle today}\\
    \bf DRIVER: & thanks\\
    \bf CAR: & \emph{you're welcome!}\\
  \end{tabular}

  \caption{Sample dialogues from our model. Note in the weather dialogue, the current day is italicized.}
\end{figure*}

\begin{figure*}
    \caption{An image provided to users to elicit unbiased audio commands for prompting more naturalistic dialogues}
    \centering
    \includegraphics[width=0.75\textwidth]{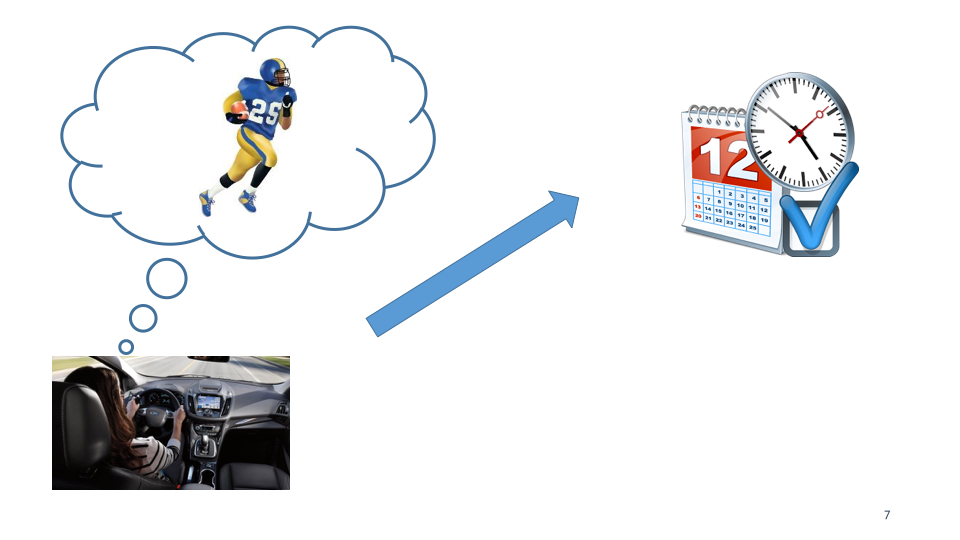}
\end{figure*} \vspace{1in}

\begin{figure*}
    \caption{\emph{Driver} mode in the wizard-of-oz collection scheme}
    \centering
    \includegraphics[width=0.75\textwidth]{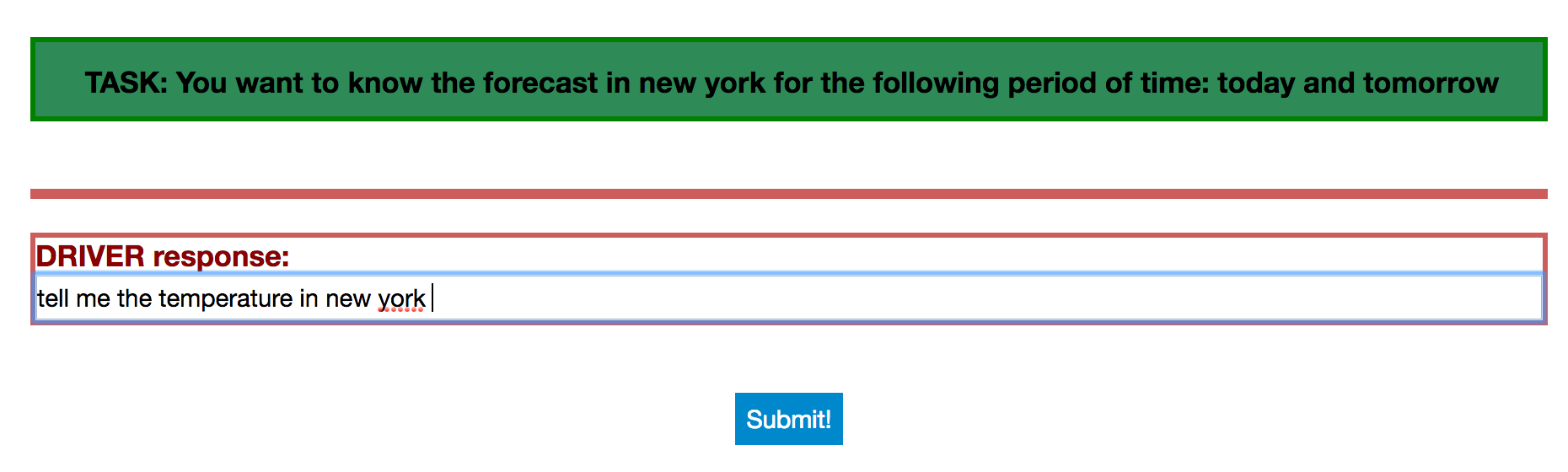} 
\end{figure*} \vspace{1in}

\begin{figure*}
    \caption{\emph{Car Assistant} mode in the wizard-of-oz collection scheme}
    \centering
    \includegraphics[width=0.5\textwidth]{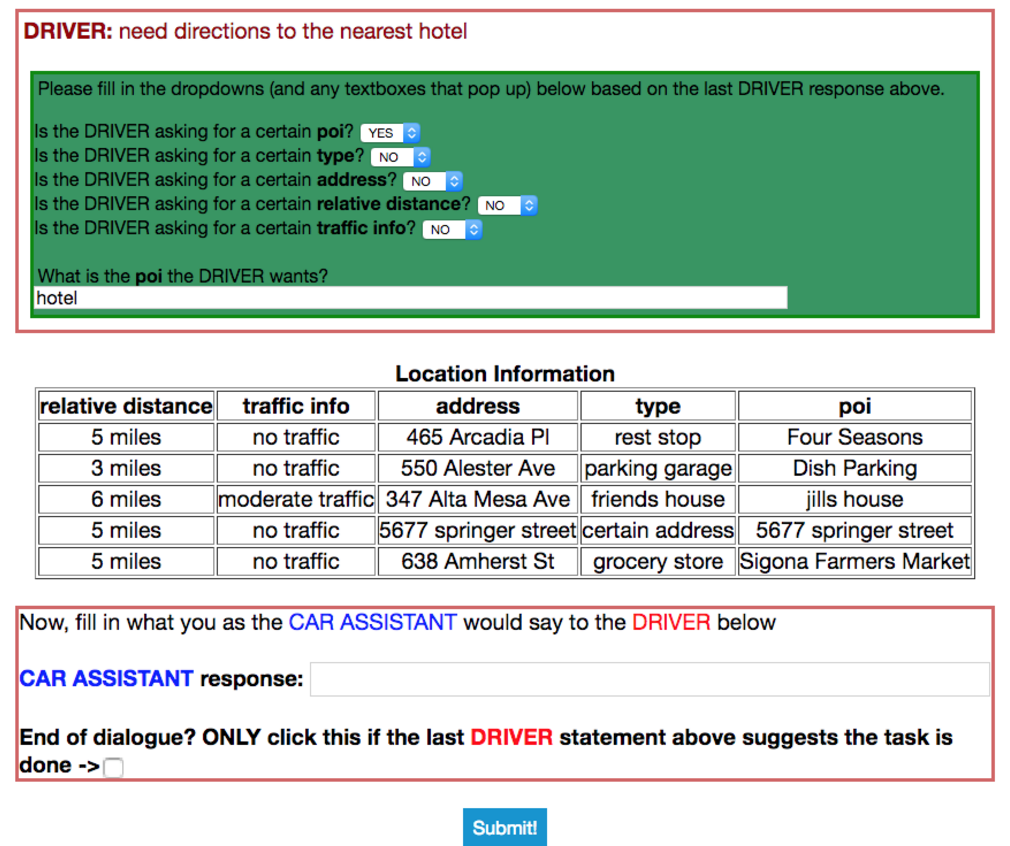}
\end{figure*}

\end{document}